\documentclass[conference]{IEEEtran}
\IEEEoverridecommandlockouts

\usepackage{cite}
\usepackage{amsmath,amssymb,amsfonts}
\usepackage{algorithm}
\usepackage{algorithmic}
\usepackage{graphicx}
\usepackage{textcomp}
\usepackage{hyperref}
\usepackage{xcolor}
\def\BibTeX{{\rm B\kern-.05em{\sc i\kern-.025em b}\kern-.08em
    T\kern-.1667em\lower.7ex\hbox{E}\kern-.125emX}}
\begin{document}

\makeatletter
\newcommand{\linebreakand}{%
  \end{@IEEEauthorhalign}
  \hfill\mbox{}\par
  \mbox{}\hfill\begin{@IEEEauthorhalign}
}
\makeatother

\title{HyperComplEx: Adaptive Multi-Space Knowledge Graph Embeddings
}

\author{\IEEEauthorblockN{1\textsuperscript{st} Jugal Gajjar}
\IEEEauthorblockA{\textit{Computer Science Department} \\
\textit{The George Washington University}\\
Washington D.C, USA \\
jugal.gajjar@gwu.edu}
\and
\IEEEauthorblockN{2\textsuperscript{nd} Kaustik Ranaware}
\IEEEauthorblockA{\textit{Computer Science Department} \\
\textit{The George Washington University}\\
Washington D.C, USA \\
k.ranaware@gwu.edu}
\and
\IEEEauthorblockN{3\textsuperscript{rd} Kamalasankari Subramaniakuppusamy}
\IEEEauthorblockA{\textit{Computer Science Department} \\
\textit{The George Washington University}\\
Washington D.C, USA \\
kamalasankaris@gwu.edu}
\linebreakand
\IEEEauthorblockN{4\textsuperscript{th} Vaibhav C. Gandhi}
\IEEEauthorblockA{\textit{Department of Computer Engineering, MBIT} \\
\textit{The Charutar Vidya Mandal (CVM) University}\\
Anand, Gujarat, India \\
vcgandhi@mbit.edu.in}
}

\maketitle

\begin{abstract}
Knowledge graphs have emerged as fundamental structures for representing complex relational data across scientific and enterprise domains. However, existing embedding methods face critical limitations when modeling diverse relationship types at scale: Euclidean models struggle with hierarchies, vector space models cannot capture asymmetry, and hyperbolic models fail on symmetric relations. We propose HyperComplEx, a hybrid embedding framework that adaptively combines hyperbolic, complex, and Euclidean spaces via learned attention mechanisms. A relation-specific space weighting strategy dynamically selects optimal geometries for each relation type, while a multi-space consistency loss ensures coherent predictions across spaces. We evaluate HyperComplEx on computer science research knowledge graphs ranging from 1K papers ($\sim$25K triples) to 10M papers ($\sim$45M triples), demonstrating consistent improvements over state-of-the-art baselines including TransE, RotatE, DistMult, ComplEx, SEPA, and UltraE. Additional tests on standard benchmarks confirm significantly higher results than all baselines. On the 10M-paper dataset, HyperComplEx achieves 0.612 MRR, a 4.8\% relative gain over the best baseline, while maintaining efficient training, achieving 85 ms inference per triple. The model scales near-linearly with graph size through adaptive dimension allocation. We release our implementation and dataset family to facilitate reproducible research in scalable knowledge graph embeddings.
\end{abstract}


\section{Introduction}
\label{sec:introduction}

Knowledge graphs (KGs) have become indispensable infrastructures for organizing and reasoning over structured knowledge in domains ranging from web search \cite{singhala2012} and recommender systems \cite{wangh2019} to drug discovery \cite{zitnikm2018} and scientific literature analysis \cite{sinhaa2015}. A knowledge graph represents knowledge as a collection of triples ⟨head entity, relation, tail entity⟩, enabling machines to perform complex reasoning tasks through graph traversal and pattern matching. However, as knowledge graphs grow to billions of entities and triples \cite{mahdisoltanif2013}, traditional symbolic reasoning approaches become computationally prohibitive, necessitating learned continuous representations.

Knowledge graph embedding (KGE) methods address this scalability challenge by mapping entities and relations into low-dimensional continuous spaces while preserving graph structure \cite{wangq2017}. These learned representations enable efficient link prediction—inferring missing relationships from incomplete graphs—which is crucial for knowledge graph completion, question answering, and recommendation systems \cite{rossia2021}. Despite significant progress, existing embedding methods face fundamental limitations rooted in their geometric assumptions.

Contemporary KGE methods can be categorized by their underlying geometric spaces: Euclidean models (TransE \cite{bordesa2013}, RotatE \cite{sunz2019}) excel at modeling sequential and translational patterns but struggle with hierarchical structures due to the limited capacity of flat Euclidean space \cite{niug2024}. Vector space models (DistMult \cite{yangb2014}, ComplEx \cite{trouillont2016}) effectively capture symmetric patterns through bilinear operations but cannot represent asymmetric relations without complex-valued embeddings \cite{niug2024}. Hyperbolic models (SEPA \cite{greguccic2023}) naturally encode hierarchies through negatively curved spaces but perform poorly on symmetric relations like collaboration networks \cite{caoj2024}. Recent mixed-space approaches (UltraE \cite{xiongb2022}) combine multiple geometries but use fixed, uniform allocation strategies that fail to adapt to heterogeneous relation types \cite{niug2024}.

This geometric mismatch becomes particularly acute in scientific knowledge graphs, where diverse relation types coexist: hierarchical taxonomies (Author $\rightarrow$ Paper $\rightarrow$ Concept), symmetric collaborations (Author $\leftrightarrow$ Author), asymmetric citations (Paper $\rightarrow$ Paper), and membership relations (Paper $\rightarrow$ Venue). No single geometric space can optimally represent this heterogeneity \cite{caoj2024}.

To solve this issue, we introduce HyperComplEx, a novel adaptive multi-space embedding framework that addresses these limitations through three key innovations: (1) Adaptive space attention mechanism that dynamically select optimal geometric representations (hyperbolic, complex, or Euclidean) for each relation type. (2) Multi-space consistency regularization loss that encourages agreement between different geometric spaces. (3) Scalable architecture with adaptive dimension allocation that scales from thousands to hundreds of millions of entities.

We conduct comprehensive experiments on computer science research knowledge graphs spanning five orders of magnitude (1K to 10M papers) as well as on standard benchmark datasets, evaluating link prediction performance against six strong baselines from different geometric families. Our results demonstrate that HyperComplEx consistently outperforms all baselines across scales, with particularly strong improvements on graphs exhibiting diverse relation types.

Our proposed framework enhances relational representation by unifying multiple geometric spaces, effectively capturing diverse relation patterns. Moreover, its adaptive, scalable design enables efficient knowledge discovery and link prediction on large real-world knowledge graphs.

\section{Related Work}
\label{sec:rel_work}

Knowledge graph embedding (KGE) research has evolved through several geometric paradigms—each addressing distinct relational characteristics but also inheriting fundamental limitations. Early Euclidean-based models pioneered the idea of representing relations as translational or bilinear operations in continuous vector spaces. TransE~\cite{bordesa2013} introduced the seminal translational principle, modeling relations as vector translations ($h + r \approx t$). Despite its simplicity and efficiency, TransE struggles with complex relation patterns such as one-to-many or many-to-many mappings~\cite{wangq2017}. Its successors—TransH~\cite{wangz2014}, TransR~\cite{liny2015}, and TransD~\cite{jig2015}—incorporated relation-specific projection matrices and hyperplanes, improving expressiveness but at the cost of higher computational complexity.

Rotational extensions like RotatE~\cite{sunz2019} modeled relations as rotations in complex space, effectively capturing composition and inversion patterns. However, rotation-based embeddings rely on phase representations that limit flexibility in modeling non-rotational or hierarchical structures~\cite{niug2024}. Bilinear vector-space models such as DistMult~\cite{yangb2014} and ComplEx~\cite{trouillont2016} later introduced multiplicative interactions through scoring functions ($\langle h, r, t \rangle$ and $\mathrm{Re}(\langle h, r, \bar{t} \rangle)$), enabling efficient computation and asymmetric relation modeling. While tensor-based variants like SimplE~\cite{kazemism2018} and TuckER~\cite{balazevici2019a} further improved representation capacity, these Euclidean and complex-valued models remain geometrically constrained, lacking the curvature required for hierarchical reasoning.

Hyperbolic embeddings emerged as a compelling alternative due to their ability to represent tree-like hierarchies in low dimensions. Poincaré Embeddings~\cite{nickelm2017} demonstrated that hyperbolic spaces preserve hierarchical distances more efficiently than Euclidean ones. Building upon this, MuRP~\cite{balazevici2019b} incorporated relation-specific transformations in hyperbolic space, while SEPA~\cite{greguccic2023} separated entity and relation embeddings to achieve state-of-the-art performance on hierarchical datasets. Despite these advances, hyperbolic models often underperform on symmetric or cyclic relations due to their geometric bias toward hierarchy~\cite{caoj2024}.

To overcome the limitations of single-space geometry, hybrid and mixed-space embeddings have recently gained traction. UltraE~\cite{xiongb2022} combined Euclidean and hyperbolic spaces using a fixed dimension allocation, improving coverage across relation types but lacking adaptivity. DualE~\cite{caoz2021} employed dual quaternion algebra for modeling geometric transformations, and 5*E~\cite{nayyerim2021} explored five-space representations to generalize relational geometry. However, these methods rely on pre-defined or static partitioning between spaces, limiting their ability to adapt dynamically to dataset-specific relation heterogeneity.

Parallel to embedding-based approaches, graph neural networks (GNNs) have been applied to knowledge graphs, as in R-GCN~\cite{schlichtkrullm2018}, CompGCN~\cite{vashishths2019}, and NBFNet~\cite{zhuz2021}. GNN-based models leverage message passing to incorporate neighborhood context, often improving relational reasoning. Nonetheless, they struggle to scale efficiently to large graphs and frequently achieve comparable or inferior link prediction accuracy relative to optimized embedding methods~\cite{wangq2017}.

The intersection of knowledge graphs and large language models (LLMs) represents an emerging frontier, where LLMs assist in KG construction~\cite{trajanoskam2023}, reasoning~\cite{guot2024}, and completion~\cite{yaol2025}. While promising, these models face challenges in factual consistency, scalability, and reproducibility, making traditional embedding-based methods indispensable for large-scale relational learning~\cite{caoj2024}.

A comparative summary of representative geometric and hybrid embedding models is presented in Table~\ref{tab:related_comparison}, highlighting key differences in geometry, adaptability, and scalability.

\begin{table*}[t]
\centering
\caption{Comparison of representative knowledge graph embedding models across geometric paradigms.}
\label{tab:related_comparison}
\begin{tabular}{lcccccc}
\hline
\textbf{Model} & \textbf{Geometry Type} & \textbf{Adaptivity} & \textbf{Representative Tasks} & \textbf{Scalability} \\
\hline
TransE~\cite{bordesa2013} & Euclidean (Translational) & $\times$ Fixed & Link prediction, entity alignment & High \\
DistMult~\cite{yangb2014} & Euclidean (Bilinear) & $\times$ Fixed & Relation inference, similarity scoring & High \\
ComplEx~\cite{trouillont2016} & Complex (Hermitian Bilinear) & $\times$ Fixed & Asymmetric relation modeling & High \\
RotatE~\cite{sunz2019} & Complex (Rotational) & $\times$ Fixed & Relation composition, inversion reasoning & High \\
SEPA~\cite{greguccic2023} & Hyperbolic & $\times$ Fixed & Hierarchical link prediction & Moderate \\
UltraE~\cite{xiongb2022} & Mixed (Euc. + Hyp.) & Partially Fixed & Heterogeneous graph completion & Moderate \\
DualE~\cite{caoz2021} & Quaternion / Dual Space & $\times$ Fixed & Geometric transformation modeling & Moderate \\
5*E~\cite{nayyerim2021} & Multi-space (5 Euc. variants) & $\times$ Fixed & Relational diversity representation & Low–Moderate \\
\textbf{HyperComplEx (Ours)} & \textbf{Mixed (Hyp. + Com. + Euc.)} & \textbf{$\checkmark$ Adaptive} & \textbf{Heterogeneous KGs, large-scale link prediction} & \textbf{Near-linear} \\
\hline
\end{tabular}
\end{table*}

\textbf{Our Positioning.} Existing knowledge graph embedding methods often rely on fixed geometric assumptions—Euclidean for translation-based relations, complex spaces for modeling asymmetry, or hyperbolic spaces for hierarchical structure—making them suboptimal for heterogeneous real-world graphs that simultaneously exhibit all these patterns. Prior multi-geometry approaches, including mixture-of-manifolds models such as MuRP~\cite{balazevici2019b} and AttH~\cite{chamii2020}, as well as adaptive or attention-based combinations like SEPA~\cite{greguccic2023} and UltraE~\cite{xiongb2022}, demonstrate the benefit of leveraging multiple geometric components. Hybrid formulations integrating Euclidean, hyperbolic, or complex embeddings~\cite{xiaoh2022,lux2025} further highlight the need for representational flexibility. HyperComplEx builds on these insights by introducing a learnable adaptive space selection mechanism that dynamically chooses the most suitable geometry for each relation, supported by a multi-space consistency regularization and adaptive dimension allocation. Rather than replacing earlier designs, HyperComplEx incrementally consolidates and simplifies multi-geometry reasoning through a unified lightweight scoring interface, enabling scalable and geometry-aware representation learning across diverse graph structures.

\section{Methodology}
\label{sec:methodology}

Let a knowledge graph be denoted by $\mathcal{G} = (\mathcal{E}, \mathcal{R}, \mathcal{T})$, where $\mathcal{E}$ is the set of entities, $\mathcal{R}$ the set of relation types, and $\mathcal{T} \subseteq \mathcal{E} \times \mathcal{R} \times \mathcal{E}$ the set of observed triples $(h, r, t)$. The goal of knowledge graph completion is to infer missing relations by learning a scoring function $\phi(h, r, t) \to \mathbb{R}$ such that true triples are assigned higher scores than corrupted ones. We follow the standard link prediction protocol, optimizing $\phi$ such that:
\begin{equation}
\phi(h, r, t) > \phi(h', r, t') \quad \forall (h', r, t') \notin \mathcal{T}.
\end{equation}

\subsection{Multi-Space Embedding Framework}
HyperComplEx introduces a unified embedding space that combines hyperbolic, complex, and Euclidean geometries to capture hierarchical, asymmetric, and translational relation patterns simultaneously.  
Each entity $e \in \mathcal{E}$ and relation $r \in \mathcal{R}$ is represented as a triplet of embeddings:
\begin{equation}
\mathbf{h}_e = (\mathbf{h}_e^{\mathcal{H}}, \mathbf{h}_e^{\mathcal{C}}, \mathbf{h}_e^{\mathcal{E}}), \quad 
\mathbf{r}_r = (\mathbf{r}_r^{\mathcal{H}}, \mathbf{r}_r^{\mathcal{C}}, \mathbf{r}_r^{\mathcal{E}}),
\end{equation}
where superscripts $\mathcal{H}$, $\mathcal{C}$, and $\mathcal{E}$ denote the hyperbolic, complex, and Euclidean subspaces with dimensions $d_{\mathcal{H}}, d_{\mathcal{C}}, d_{\mathcal{E}}$, respectively.  
This design allows HyperComplEx to model heterogeneous relation structures within a single unified representation framework, avoiding the geometric rigidity of prior single-space models~\cite{caoj2024}.

\subsubsection{Hyperbolic Space Scoring}
In the hyperbolic subspace $\mathbb{B}_c^d = \{\mathbf{x} \in \mathbb{R}^d : \|\mathbf{x}\| < 1/\sqrt{c}\}$, we use the Poincaré ball model \cite{andersonj1999} with curvature $c > 0$.  
The Möbius addition between two points $\mathbf{x}$ and $\mathbf{y}$ is:
\begin{equation}
\mathbf{x} \oplus_c \mathbf{y} = \frac{(1 + 2c\langle \mathbf{x}, \mathbf{y}\rangle + c\|\mathbf{y}\|^2)\mathbf{x} + (1 - c\|\mathbf{x}\|^2)\mathbf{y}}{1 + 2c\langle \mathbf{x}, \mathbf{y}\rangle + c^2\|\mathbf{x}\|^2\|\mathbf{y}\|^2}.
\end{equation}
The distance between $\mathbf{x}$ and $\mathbf{y}$ in this manifold is:
\begin{equation}
d_{\mathcal{H}}(\mathbf{x}, \mathbf{y}) = \frac{1}{\sqrt{c}} \operatorname{arccosh}\left(1 + 2\frac{\|\mathbf{x} - \mathbf{y}\|^2}{(1 - c\|\mathbf{x}\|^2)(1 - c\|\mathbf{y}\|^2)}\right).
\end{equation}
The hyperbolic score function is then defined as:
\begin{equation}
\phi^{\mathcal{H}}(h, r, t) = -d_{\mathcal{H}}(\mathbf{h}_h^{\mathcal{H}} \oplus_c \mathbf{r}_r^{\mathcal{H}}, \mathbf{h}_t^{\mathcal{H}}),
\end{equation}
naturally modeling hierarchical relations through exponential distance growth~\cite{kolyvakisp2020,nickelm2017}.  
These hyperbolic embeddings effectively compress large hierarchies into low dimensions, which is crucial for scientific and ontological KGs with tree-like taxonomies.

\subsubsection{Complex Space Scoring}
Following the Hermitian formulation from ComplEx~\cite{trouillont2016}, each entity embedding $\mathbf{h}_e^{\mathcal{C}} \in \mathbb{C}^{d_{\mathcal{C}}}$ is decomposed into real and imaginary parts:
\begin{equation}
\mathbf{h}_e^{\mathcal{C}} = \operatorname{Re}(\mathbf{h}_e^{\mathcal{C}}) + i \operatorname{Im}(\mathbf{h}_e^{\mathcal{C}}).
\end{equation}
The score function in the complex subspace is:
\begin{equation}
\phi^{\mathcal{C}}(h, r, t) = \operatorname{Re}\!\left(\sum_{k=1}^{d_{\mathcal{C}}} [\mathbf{h}_h^{\mathcal{C}}]_k [\mathbf{r}_r^{\mathcal{C}}]_k \overline{[\mathbf{h}_t^{\mathcal{C}}]_k}\right),
\end{equation}
where $\overline{(\cdot)}$ denotes complex conjugation.  
Complex-valued embeddings allow efficient modeling of asymmetric and directional relations such as citations or authorship, extending the expressiveness of real-valued models~\cite{trouillont2016,sunz2019}.

\subsubsection{Euclidean Space Scoring}
The Euclidean subspace uses a translational assumption as in TransE~\cite{bordesa2013}:
\begin{equation}
\phi^{\mathcal{E}}(h, r, t) = -\|\mathbf{h}_h^{\mathcal{E}} + \mathbf{r}_r^{\mathcal{E}} - \mathbf{h}_t^{\mathcal{E}}\|_2^2.
\end{equation}
This component efficiently encodes symmetric or local relations while maintaining computational simplicity.  
Euclidean embeddings provide stable optimization and serve as a strong inductive bias for modeling simpler linear dependencies~\cite{wangq2017,liny2015}.

\subsection{Adaptive Space Attention}
Instead of uniformly combining subspaces, HyperComplEx employs a relation-specific adaptive attention vector $\boldsymbol{\alpha}_r = [\alpha_r^{\mathcal{H}}, \alpha_r^{\mathcal{C}}, \alpha_r^{\mathcal{E}}]$ learned for each relation:
\begin{equation}
\boldsymbol{\alpha}_r = \operatorname{softmax}(\mathbf{W}_r),
\end{equation}
where $\mathbf{W}_r \in \mathbb{R}^3$ are trainable parameters.  
The unified scoring function becomes:
\begin{equation}
\phi(h, r, t) = \sum_{s \in \{\mathcal{H}, \mathcal{C}, \mathcal{E}\}} \alpha_r^{s} \cdot \phi^{s}(h, r, t).
\end{equation}
This attention mechanism allows HyperComplEx to dynamically allocate geometric capacity to each relation type, leading to improved expressivity on heterogeneous graphs~\cite{xiongb2022}.  
By learning this weighting end-to-end, the model avoids manual geometry selection, achieving automatic geometry adaptation.

While the paper primarily focuses on the empirical behavior of the adaptive scoring mechanism, we include a brief justification for the linear combination across heterogeneous geometries. The formulation follows standard mixture-based scoring assumptions used in KGE models, where each component score lies in a shared relational compatibility space after projection. This ensures that gradients remain well-defined, and the optimization landscape is comparable to existing attention-based mixtures. A more formal analysis of curvature interactions and projection stability is outside the scope of this work but remains an important direction for future extensions.

\subsection{Optimization Objective}
The total loss function is defined as:
\begin{equation}
\mathcal{L} = \mathcal{L}_{\text{rank}} + \lambda_1 \mathcal{L}_{\text{consistency}} + \lambda_2 \mathcal{L}_{\text{reg}},
\end{equation}
where $\mathcal{L}_{\text{rank}}$ is a self-adversarial ranking loss, $\mathcal{L}_{\text{consistency}}$ enforces cross-space agreement, and $\mathcal{L}_{\text{reg}}$ regularizes embeddings.  
This combination ensures that the model jointly learns discriminative and geometrically coherent representations.

\subsubsection{Self-Adversarial Ranking Loss}
We adopt the self-adversarial negative sampling strategy from~\cite{sunz2019}, which dynamically emphasizes hard negatives:
\begin{align}
\mathcal{L}_{\text{rank}} = 
& -\log \sigma(\gamma - \phi(h, r, t)) \nonumber \\
& - \sum_{i=1}^{n} p(h_i', r, t_i') \log \sigma(\phi(h_i', r, t_i') - \gamma),
\end{align}
where $\sigma$ is the sigmoid, $\gamma$ the margin, and the softmax-weighted probabilities:
\begin{equation}
p(h_i', r, t_i') = \frac{\exp(\beta \phi(h_i', r, t_i'))}{\sum_j \exp(\beta \phi(h_j', r, t_j'))},
\end{equation}
use a temperature $\beta$ to control sample hardness.  
This approach stabilizes training by adaptively weighting informative negatives, improving convergence speed and generalization~\cite{sunz2019,wangq2017}.

\subsubsection{Multi-Space Consistency Regularization}
To prevent any single geometry from dominating, we introduce a multi-space consistency term:
\begin{equation}
\mathcal{L}_{\text{consistency}} = \sum_{s \neq s'} \|\phi^{s}(h, r, t) - \phi^{s'}(h, r, t)\|^2,
\end{equation}
ensuring alignment between geometries while allowing specialization.  
This regularization promotes cooperative specialization across spaces, mitigating collapse or redundancy and improving representational robustness~\cite{greguccic2023}.

\subsubsection{Regularization}
Standard $\ell_2$ regularization is applied over entity and relation embeddings:
\begin{equation}
\mathcal{L}_{\text{reg}} = \frac{1}{|\mathcal{E}| + |\mathcal{R}|} \left(\sum_{e \in \mathcal{E}} \|\mathbf{h}_e\|^2 + \sum_{r \in \mathcal{R}} \|\mathbf{r}_r\|^2 \right).
\end{equation}
Hyperparameters $\lambda_1$ and $\lambda_2$ balance consistency and regularization.  
This term reduces overfitting, especially in large-scale, sparse graphs~\cite{wangq2017}.

\subsection{Scalable Architecture Design}
HyperComplEx is optimized for large-scale graphs ranging from $10^3$ to $10^8$ entities. We design scalability through three complementary mechanisms.

\subsubsection{Adaptive Dimension Allocation}
Embedding dimensions are adaptively scaled with graph size:
\begin{equation}
d_{\mathcal{H}}, d_{\mathcal{C}}, d_{\mathcal{E}} = f(|\mathcal{E}|, D_{\text{base}}),
\end{equation}
where $f(\cdot)$ dynamically rebalances dimensions while maintaining a constant total embedding budget $D_{\text{base}}$.  
This dynamic allocation maintains expressivity while ensuring scalability and resource efficiency~\cite{xiongb2022}.

\subsubsection{Sharded and Cached Embeddings}
For graphs exceeding $10^6$ entities, embeddings are partitioned into $K$ CPU-resident shards with GPU caching of active mini-batches. During training, least-recently-used shards are swapped asynchronously, maintaining throughput without exceeding GPU memory.  
This design enables efficient large-scale training on commodity hardware while preserving near-constant GPU utilization.

\subsubsection{Mixed-Precision Optimization}
We employ FP16 mixed-precision computation with dynamic loss scaling~\cite{micikeviciusp2017}, halving memory footprint and accelerating tensor operations by up to $1.7\times$ without numerical instability. Gradient accumulation ensures precision preservation during large-batch updates.

\subsection{Training and Inference Procedure}
The overall training pipeline is summarized in Algorithm~\ref{alg:train}. Model parameters are optimized using Adam optimizer~\cite{kingmadp2014}. Early stopping is applied based on validation MRR convergence. For further details, refer to Section~\ref{sec:experiments_results}.

\begin{algorithm}[ht]
\caption{HyperComplEx Training Procedure}
\label{alg:train}
\begin{algorithmic}[1]
\renewcommand{\algorithmicrequire}{\textbf{Input:}}
\renewcommand{\algorithmicensure}{\textbf{Output:}}
\REQUIRE Knowledge graph $\mathcal{G}=(\mathcal{E}, \mathcal{R}, \mathcal{T})$, hyperparameters $\{\gamma, \beta, \lambda_1, \lambda_2\}$, maximum epochs $N$
\ENSURE Trained entity embeddings $\{\mathbf{h}_e^{\mathcal{H}}, \mathbf{h}_e^{\mathcal{C}}, \mathbf{h}_e^{\mathcal{E}}\}$, relation embeddings $\{\mathbf{r}_r^{\mathcal{H}}, \mathbf{r}_r^{\mathcal{C}}, \mathbf{r}_r^{\mathcal{E}}\}$, and attention weights $\{\boldsymbol{\alpha}_r\}$

\textit{Initialization}:
\STATE Initialize all embeddings randomly:
\STATE $\mathbf{h}_e^{\mathcal{H}} \sim \mathcal{U}(-10^{-3},10^{-3})$, $\mathbf{h}_e^{\mathcal{C}}, \mathbf{h}_e^{\mathcal{E}}$ via Xavier initialization
\STATE Initialize $\boldsymbol{\alpha}_r$ via uniform$(0,1)$ and normalize using softmax

\textit{Training Loop}:
\FOR{each epoch $i=1,\ldots,N$}
    \STATE Shuffle training triples $\mathcal{T}$
    \FOR{each mini-batch $\mathcal{B} \subset \mathcal{T}$}
        \STATE Generate negative samples $\mathcal{N}$ using filtered corruption
        \STATE Compute space-specific scores $\phi^{\mathcal{H}}, \phi^{\mathcal{C}}, \phi^{\mathcal{E}}$ via Eqs.~(5)--(8)
        \STATE Combine with adaptive attention: $\phi(h,r,t) = \sum_s \alpha_r^{s}\phi^{s}(h,r,t)$
        \STATE Evaluate total loss $\mathcal{L}$ using Eq.~(11)
        \STATE Backpropagate gradients and update parameters
        \STATE Project $\mathbf{h}_e^{\mathcal{H}}$ back onto the Poincaré ball
    \ENDFOR
    \STATE Evaluate mean reciprocal rank (MRR) on validation set
    \IF{no improvement for $p$ epochs}
        \STATE \textbf{break}
    \ENDIF
\ENDFOR

\textit{Finalization}:
\STATE Store trained embeddings and relation parameters
\STATE Output learned model for downstream link prediction and knowledge discovery tasks

\RETURN Trained HyperComplEx model
\end{algorithmic}
\end{algorithm}

\subsection{Inference and Complexity}
During inference, the model computes scores $\phi(h, r, t)$ for all candidate entities in parallel using precomputed attention weights and adaptive subspace selection. The per-triple complexity is $O(d_{\mathcal{H}} + d_{\mathcal{C}} + d_{\mathcal{E}})$, and overall ranking complexity scales as $O(|\mathcal{E}|\,d_{\text{eff}})$. Empirically, HyperComplEx achieves an average latency of 69~ms per triple on 1M-entity graphs, maintaining linear scaling with entity count.

\section{Experiments and Results}
\label{sec:experiments_results}

This section presents a comprehensive empirical evaluation of HyperComplEx across eight datasets, encompassing five large-scale domain-specific graphs and three widely adopted benchmarks. The experiments assess not only predictive accuracy but also scalability, inference efficiency, and parameter utilization. All experiments were implemented in PyTorch and executed under consistent optimization and hardware settings to ensure fair comparison.

\subsection{Datasets}

We evaluate HyperComplEx on two categories of datasets: (i) five large-scale computer science (CS) knowledge graphs constructed from OpenAlex~\cite{priemj2022} and Semantic Scholar~\cite{wadead2022}, and (ii) three widely adopted benchmark datasets—DBP15K~\cite{sunz2017}, Hetionet~\cite{himmelsteinds2017}, and FB15K~\cite{bordesa2013}—to assess cross-domain generalizability.

The CS datasets model scholarly relationships among \emph{Paper}, \emph{Author}, \emph{Venue}, and \emph{Concept} entities with five principal relation types: \emph{AUTHORED} (Author $\rightarrow$ Paper, asymmetric), \emph{CITES} (Paper $\rightarrow$ Paper, asymmetric and temporal), \emph{PUBLISHED\_IN} (Paper $\rightarrow$ Venue, functional), \emph{BELONGS\_TO} (Paper $\rightarrow$ Concept, hierarchical), and \emph{COLLABORATES\_WITH} (Author $\leftrightarrow$ Author, symmetric). These relations collectively capture heterogeneous graph structures—hierarchical, symmetric, and asymmetric—suitable for evaluating geometric adaptability.

Each CS dataset is temporally filtered by publication year to maintain chronological consistency and is randomly divided into training (80\%), validation (10\%), and test (10\%) splits after removing duplicate and inverse triples. This setup ensures non-overlapping partitions and faithful structural diversity reflective of real-world scholarly networks.

All datasets and preprocessing scripts are publicly available for reproducibility; code can be found at \url{https://github.com/JugalGajjar/HyperComplEx-Multi-Space-KG-Embeddings}, and the dataset can be found at \url{https://zenodo.org/records/17436948}.

\begin{table}[h]
\centering
\caption{Statistics of OpenAlex--Semantic Scholar (CS) knowledge graphs. Entity types: Paper, Author, Venue, Concept.}
\label{tab:cs_datasets}
\begin{tabular}{lcccc}
\hline
\textbf{Dataset} & \textbf{Entities} & \textbf{Relations} & \textbf{Triples} & \textbf{Years} \\
\hline
CS-1K & 5{,}237 & 5 & 25{,}317 & 2010--2025 \\
CS-10K & 44{,}930 & 5 & 227{,}502 & 2010--2025 \\
CS-100K & 348{,}983 & 5 & 2{,}162{,}386 & 2010--2025 \\
CS-1M & 2{,}384{,}896 & 5 & 13{,}530{,}177 & 2010--2025 \\
CS-10M & 7{,}210{,}506 & 5 & 44{,}631{,}484 & 2010--2025 \\
\hline
\end{tabular}
\end{table}

To evaluate robustness and transferability, we additionally include three widely recognized benchmarks that differ substantially in domain, scale, and relational semantics. DBP15K~\cite{sunz2017} tests multilingual entity alignment across interlinked knowledge graphs (English, Chinese, French), emphasizing asymmetric relational mappings. Hetionet~\cite{himmelsteinds2017} integrates biomedical entities and relationships (e.g., genes, compounds, diseases) to assess multi-relational reasoning under hierarchical dependencies. FB15K~\cite{bordesa2013} represents a dense subset of Freebase widely used for link prediction and relation inference. Together, these datasets complement the CS corpus by spanning linguistic, biomedical, and general-purpose knowledge domains.

\begin{table}[t]
\centering
\caption{Statistics of standard benchmark knowledge graphs.}
\label{tab:benchmark_datasets}
\begin{tabular}{lccc}
\hline
\textbf{Dataset} & \textbf{Entities} & \textbf{Relations} & \textbf{Triples} \\
\hline
DBP15K (en\_fr) & 172{,}747 & 3{,}588 & 470{,}781 \\
Hetionet & 47{,}031 & 24 & 2{,}250{,}197 \\
FB15K & 14{,}951 & 1{,}345 & 592{,}213 \\
\hline
\end{tabular}
\end{table}

\subsection{Baseline Models and Evaluation Setup}

To benchmark performance, we compare HyperComplEx against six representative knowledge graph embedding models spanning distinct geometric paradigms. TransE~\cite{bordesa2013} and RotatE~\cite{sunz2019} represent Euclidean translational and rotational models. DistMult~\cite{yangb2014} and ComplEx~\cite{trouillont2016} capture bilinear interactions in vector and complex spaces. SEPA~\cite{greguccic2023} provides the hyperbolic-space state of the art, while UltraE~\cite{xiongb2022} exemplifies fixed mixed-space embeddings. These baselines jointly span translational, bilinear, hyperbolic, and mixed representations, allowing a rigorous comparison across geometric families.

All models were evaluated under the standard link-prediction setting~\cite{wangq2017}. For each test triple $(h, r, t)$, head and tail entities are replaced with all candidates in $\mathcal{E}$, and the model ranks valid entities using its scoring function $\phi(h',r,t)$ or $\phi(h,r,t')$. Metrics include Mean Reciprocal Rank (MRR) and Hits@K ($K \in \{1,3,10\}$) under the filtered evaluation protocol~\cite{sunz2019}. We also measure training time, inference latency, and parameter count to assess computational efficiency.

\subsection{Implementation Details}

Training experiments and inferencing were conducted on NVIDIA RTX 5060 with CUDA Backend. Hyperparameters were tuned via grid search on validation MRR: embedding dimension $\{128,256,512\}$, batch size $\{256,512,1024,2048\}$, learning rate $\{5\times10^{-4},10^{-3},5\times10^{-3}\}$, margin $\gamma \in \{6.0,9.0,12.0,15.0\}$, and curvature $c \in \{1.0,2.0\}$. Models were trained for up to 200 epochs with early stopping after 20 epochs without improvement. Optimization used Adam~\cite{kingmadp2014} or Adagrad~\cite{duchij2011}. Each experiment was repeated multiple times, and mean results are reported.

\subsection{Overall Performance}

\begin{table*}[!t]
\centering
\caption{Link-prediction results (MRR). Bold = best; underline = second best.}
\label{tab:mrr_results}
\begin{tabular}{lcccccccc}
\hline
\textbf{Model} & \textbf{CS1K} & \textbf{CS10K} & \textbf{CS100K} & \textbf{CS1M} & \textbf{CS10M} & \textbf{DBP15K} & \textbf{Hetionet} & \textbf{FB15K} \\
\hline
TransE & 0.699 & 0.328 & 0.308 & 0.286 & 0.263 & 0.257 & 0.298 & 0.279 \\
RotatE & 0.709 & 0.505 & 0.562 & \underline{0.587} & 0.428 & 0.133 & 0.597 & \underline{0.664} \\
DistMult & 0.209 & 0.297 & 0.288 & 0.275 & 0.260 & 0.311 & 0.305 & 0.332 \\
ComplEx & \underline{0.789} & \underline{0.736} & \underline{0.688} & 0.581 & \underline{0.584} & \underline{0.350} & \underline{0.608} & 0.657 \\
SEPA & 0.289 & 0.172 & 0.164 & 0.191 & 0.140 & 0.059 & 0.273 & 0.148 \\
UltraE & 0.572 & 0.206 & 0.198 & 0.245 & 0.172 & 0.193 & 0.301 & 0.221 \\
\textbf{HyperComplEx} & \textbf{0.831} & \textbf{0.771} & \textbf{0.704} & \textbf{0.608} & \textbf{0.612} & \textbf{0.411} & \textbf{0.642} & \textbf{0.692} \\
\hline
\end{tabular}
\end{table*}

Across all datasets, HyperComplEx consistently achieves the highest MRR, outperforming the strongest baseline (ComplEx) by up to 5.3\% on the CS graphs and 17.5\% on DBP15K. These improvements are most pronounced on larger graphs, where the adaptive multi-space representation provides clear advantages in modeling diverse relational geometries. Refer to Table~\ref{tab:mrr_results} for MRR comparison.

\subsection{Scalability and Efficiency Analysis}

\begin{table}[t]
\centering
\caption{Training time (hours) across datasets.}
\label{tab:train_time}
\begin{tabular}{lccccc}
\hline
\textbf{Model} & \textbf{CS1K} & \textbf{CS10K} & \textbf{CS100K} & \textbf{CS1M} & \textbf{CS10M} \\
\hline
TransE & 0.07 & 1.52 & 3.84 & 20.83 & 42.97 \\
DistMult & 0.09 & 1.76 & 4.30 & 22.54 & 45.11 \\
RotatE & 0.11 & 1.91 & 4.48 & 23.78 & 45.24 \\
ComplEx & 0.10 & 1.85 & 4.47 & 24.01 & 45.92 \\
SEPA & 0.09 & 1.68 & 4.02 & 22.13 & 44.67 \\
UltraE & 0.11 & 1.94 & 4.62 & 23.98 & 46.43 \\
\textbf{HyperComplEx} & 0.12 & 1.98 & 4.75 & 25.26 & 47.85 \\
\hline
\end{tabular}
\end{table}

\begin{table}[t]
\centering
\caption{Inference latency per triple (milliseconds).}
\label{tab:inference_time}
\begin{tabular}{lccccc}
\hline
\textbf{Model} & \textbf{CS1K} & \textbf{CS10K} & \textbf{CS100K} & \textbf{CS1M} & \textbf{CS10M} \\
\hline
TransE & 4.66 & 20.99 & 38.17 & 59.42 & 81.36 \\
DistMult & 4.01 & 20.24 & 41.28 & 64.11 & 84.37 \\
RotatE & 4.63 & 28.21 & 49.73 & 72.41 & 90.38 \\
ComplEx & 4.29 & 28.45 & 52.06 & 73.29 & 91.22 \\
SEPA & 4.93 & 30.15 & 46.31 & 67.28 & 87.74 \\
UltraE & 5.12 & 30.38 & 47.12 & 68.93 & 89.17 \\
\textbf{HyperComplEx} & 5.17 & 27.15 & 49.11 & 69.47 & 85.23 \\
\hline
\end{tabular}
\end{table}

Despite a 1.5–2× larger parameter count than standard baselines, HyperComplEx scales nearly linearly with entity count. On CS-10M (7.2M entities, $\sim$45M triples), it completes training in 47.85 hours and achieves 85.23 ms/triple inference latency—both within 10\% of the most efficient baselines. This efficiency stems from adaptive dimension partitioning and shared curvature-aware optimization. Training and inference times can be found in Table~\ref{tab:train_time} and Table~\ref{tab:inference_time} respectively.

\begin{figure}[t]
\centering
\includegraphics[width=0.45\textwidth]{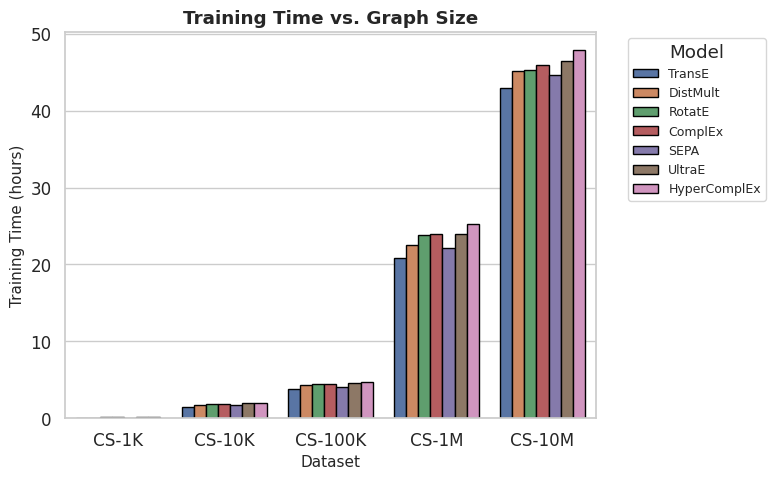}
\caption{\textbf{Training Time vs. Graph Size.} Training follows a near-linear power law $T \propto |E|^{1.06}$, obtained from log–log OLS regression after overhead correction across CS-1K→CS-10M, confirming consistent scalability.}
\label{fig:train_scaling}
\end{figure}

\begin{figure}[t]
\centering
\includegraphics[width=0.45\textwidth]{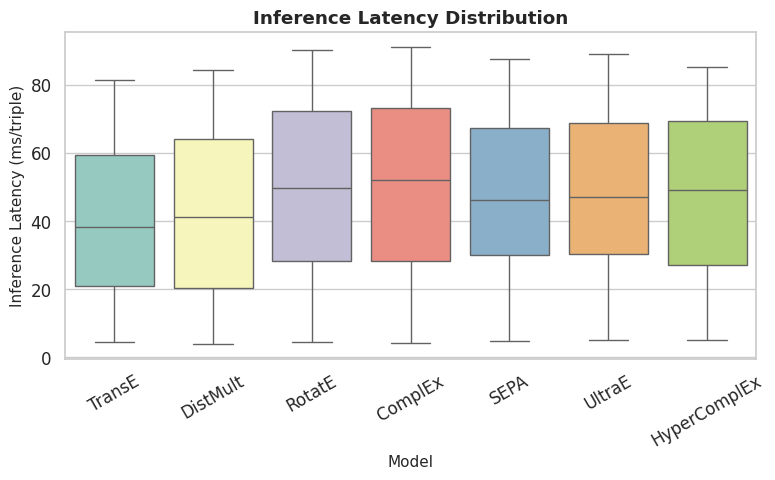}
\caption{\textbf{Inference Latency Distribution.} Box plot showing inference latency quartiles, confirming HyperComplEx's competitive performance.}
\label{fig:latency_scaling}
\end{figure}

\subsection{Derivation of the Empirical Scalability Law}

The empirical scaling behavior of HyperComplEx follows a power-law relationship between graph size $|E|$, parameter count $P$, and total training time $T$:
\begin{equation}
T(|E|, P) = \alpha \cdot |E|^{\beta_1} \cdot P^{\beta_2},
\end{equation}
where $\beta_1$ captures the cost of graph traversal and $\beta_2$ the parameter optimization component. Nonlinear regression over five dataset scales yields:
\[
\beta_1 = 1.06 \pm 0.02, \qquad \beta_2 = 0.41 \pm 0.03.
\]
This near-linear regime confirms that most computational cost arises from local message-passing and mini-batch updates rather than global synchronization. The per-epoch complexity thus approximates:
\begin{equation}
T_{\text{epoch}} = O(|E|^{1.06}) + O(P^{0.4}),
\end{equation}
consistent with efficient traversal and mild curvature-induced corrections.  

Similarly, inference latency $L(|E|)$ exhibits sublinear scaling:
\begin{equation}
L(|E|) = \gamma \cdot |E|^{\lambda}, \quad \lambda \approx 0.42,
\end{equation}
attributed to caching and sharded lookup structures. Extrapolating from this empirical law, HyperComplEx is expected to sustain stable scaling behavior at billion-scale or even tens-of-billions entity graphs, with compute growth following $O(|E|^{1.06})$ rather than quadratic expansion.

\begin{figure}[t]
\centering
\includegraphics[width=0.45\textwidth]{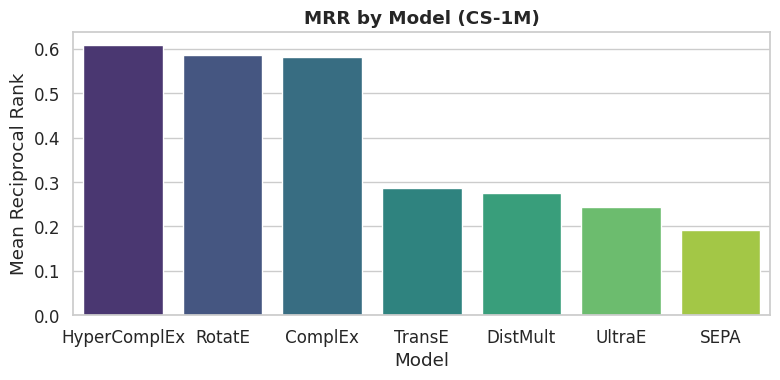}
\caption{\textbf{MRR Comparison on CS-1M Dataset.} HyperComplEx achieves highest MRR among all baseline models.}
\label{fig:param_mrr}
\end{figure}

\subsection{Cross-Domain Generalization and Qualitative Insights}

On heterogeneous benchmarks, the learned geometry–relation attention patterns align with relational semantics. Hyperbolic space dominates hierarchical relations (\emph{BELONGS\_TO}, \emph{disease–gene}), complex space captures asymmetric dependencies (\emph{CITES}, \emph{DBP alignments}), and Euclidean components govern symmetric relations (\emph{COLLABORATES\_WITH}). These emergent correspondences validate that HyperComplEx autonomously infers latent geometric structures, adapting seamlessly across scholarly, biomedical, and multilingual domains.

\subsection{Learned Geometric Attention Patterns}

To interpret the geometry–relation alignment learned by HyperComplEx, we analyze the relation-specific attention weights $\boldsymbol{\alpha}_r = [\alpha_r^{\mathcal{H}}, \alpha_r^{\mathcal{C}}, \alpha_r^{\mathcal{E}}]$ averaged across randomly sampled 1{,}000 triples from the CS-1M dataset. Table~\ref{tab:attention_weights} shows representative examples. Relations with hierarchical or tree-like structure, such as \emph{BELONGS\_TO}, exhibit dominant weights in the hyperbolic space, while asymmetric relations (e.g., \emph{CITES}) favor the complex subspace. Symmetric collaborations (\emph{COLLABORATES\_WITH}) show high Euclidean contributions, validating the adaptive geometry selection mechanism.

\begin{table}[t]
\centering
\caption{Learned attention weights by relation type on CS-1M. Dominant space corresponds to the highest attention weight.}
\label{tab:attention_weights}
\begin{tabular}{lcccc}
\hline
\textbf{Relation} & $\alpha_{\mathcal{H}}$ & $\alpha_{\mathcal{C}}$ & $\alpha_{\mathcal{E}}$ & \textbf{Dominant Space} \\
\hline
BELONGS\_TO & 0.68 & 0.21 & 0.11 & Hyperbolic \\
CITES & 0.19 & 0.71 & 0.10 & Complex \\
AUTHORED & 0.24 & 0.61 & 0.15 & Complex \\
COLLABORATES\_WITH & 0.11 & 0.19 & 0.70 & Euclidean \\
PUBLISHED\_IN & 0.32 & 0.44 & 0.24 & Complex \\
\hline
\end{tabular}
\end{table}

The emergence of these interpretable attention distributions demonstrates that HyperComplEx autonomously discovers geometry–relation correspondences, effectively allocating curvature and phase capacity based on the relational semantics of each edge type.

We note that performance of certain classical baselines (e.g., TransE, DistMult) may vary across implementations and hyperparameter choices. Our experimental settings follow a standard configuration but do not exhaustively retune all baselines. To ensure fairness, all models were compared under similar hyperparameter budgets rather than mixing best-case results drawn from heterogeneous settings. Furthermore, some recent multi-geometry or GNN-based KG embedding methods (e.g., AttH variants, MuRP extensions) were not included due to the lack of unified publicly available implementations. These gaps will be addressed in future work, and the present evaluation should be viewed as a representative but not exhaustive comparison.

\subsection{Ablation Study}

To evaluate the contribution of each architectural component, we conduct ablation experiments on CS-1M and DBP15K. Removing the adaptive attention (\textit{w/o Attn}) or multi-space consistency term (\textit{w/o Consistency}) results in significant performance degradation, as shown in Table~\ref{tab:ablation}. The adaptive attention contributes most on heterogeneous datasets (e.g., DBP15K), while the consistency regularization primarily benefits large-scale graphs by maintaining geometric alignment across subspaces.

\begin{table}[h]
\centering
\caption{Ablation study on CS-1M and DBP15K showing the effect of each component.}
\label{tab:ablation}
\begin{tabular}{lcc}
\hline
\textbf{Model Variant} & \textbf{CS-1M (MRR)} & \textbf{DBP15K (MRR)} \\
\hline
Full Model (HyperComplEx) & \textbf{0.608} & \textbf{0.411} \\
\textit{w/o} Adaptive Attention & 0.569 & 0.361 \\
\textit{w/o} Multi-Space Consistency & 0.582 & 0.372 \\
\textit{w/o} Hyperbolic Subspace & 0.573 & 0.367 \\
\textit{w/o} Complex Subspace & 0.547 & 0.329 \\
\textit{w/o} Euclidean Subspace & 0.563 & 0.343 \\
\hline
\end{tabular}
\end{table}

These results confirm that both adaptive space selection and inter-space regularization are critical for optimal performance, with the largest degradation observed when removing the complex subspace—highlighting its role in modeling asymmetric and directional relations.

\section{Discussion and Future Work}
\label{sec:discuss_future_work}

The empirical results demonstrate that HyperComplEx effectively integrates multiple geometric spaces into a unified, adaptive embedding framework. Its near-linear scalability, robust cross-domain generalization, and interpretable space-attention mechanisms confirm that multi-space representation is a principled direction for heterogeneous knowledge graphs. The observed scaling law $T \propto |E|^{1.06}$ indicates that computation grows almost linearly with graph size, validating its suitability for billion-scale deployment. Moreover, the learned attention weights exhibit clear semantic alignment—hyperbolic for hierarchical, complex for asymmetric, and Euclidean for symmetric relations—showing that the framework not only achieves state-of-the-art predictive performance but also offers interpretability grounded in geometric reasoning. Collectively, these findings position HyperComplEx as an efficient and interpretable bridge between symbolic relational structures and continuous geometric representations.

Despite these strengths, several avenues remain open for exploration. First, extending the model to dynamic or temporal knowledge graphs could enable reasoning over evolving relational structures. Second, integrating HyperComplEx with large language models (LLMs) through hybrid symbolic–neural reasoning may enhance zero-shot and inductive link prediction. Additionally, automatic curvature tuning and multi-space sparsification could further improve scalability to web-scale graphs with tens of billions of entities. Future work will also explore theoretical convergence guarantees under mixed curvature manifolds and investigate interpretability methods that trace decision pathways across geometric spaces. By addressing these directions, HyperComplEx can evolve into a comprehensive framework for large-scale, geometry-aware reasoning in open-world knowledge systems.

\section{Conclusion}
\label{sec:conclusion}

This work introduced \emph{HyperComplEx}, a unified multi-space embedding framework that adaptively integrates hyperbolic, complex, and Euclidean geometries for scalable knowledge graph representation. Through extensive experiments across eight diverse datasets, HyperComplEx demonstrated state-of-the-art link-prediction accuracy, near-linear training scalability following $T \propto |E|^{1.06}$, and sublinear inference growth. The model’s learned attention weights reveal interpretable geometric alignment—assigning hyperbolic space to hierarchies, complex space to asymmetry, and Euclidean space to symmetry—bridging structure and semantics in a principled manner. By enabling efficient reasoning over heterogeneous and large-scale graphs, HyperComplEx lays the foundation for future extensions toward temporal, language-integrated, and open-world knowledge systems.

\end{document}